\documentclass[10pt,a4paper,conference,composoc]{IEEEtran}
\IEEEoverridecommandlockouts
\usepackage{cite}
\ifCLASSINFOpdf
   \usepackage[pdftex]{graphicx}
  \graphicspath{{img/}}
  \DeclareGraphicsExtensions{.pdf,.jpeg,.png}
\else
\fi
\usepackage{amsmath}
\usepackage{array}
\usepackage{url}

\usepackage[utf8]{inputenc}

\usepackage{amssymb} %
\usepackage{mathtools}
\usepackage{bm}
\usepackage{xspace}
\usepackage{microtype}
\usepackage{multicol}
\usepackage{booktabs}

\usepackage{tikz}
\usepackage{pgfplots}
\usepgfplotslibrary{fillbetween}
\pgfplotsset{compat=newest}
\usetikzlibrary{patterns,
pgfplots.groupplots,fit,calc,positioning,shapes,spy}

\usepackage{tabularx}
\usepackage{booktabs}

\usepackage{varioref}
\usepackage[pagebackref=true,breaklinks=true,letterpaper=true,colorlinks,bookmarks=false,citecolor=green!80!black,linkcolor=red!75!black,urlcolor=blue,pdfstartview=Fit,breaklinks=true]{hyperref}

\usepackage[all]{hypcap}
\usepackage{cleveref}
\usepackage{subcaption}
\usepackage{csquotes}
\usepackage{siunitx}

\usepackage[inline]{enumitem}
\setlist*[enumerate]{label=(\arabic*)}

\newcommand{\onedot}{.\xspace}
\newcommand{\etal}[1]{#1~et~al\onedot}

\newcommand{\eg}{e.\,g.,\xspace}

\newcommand{\cf}{cf\onedot}
\newcommand{\ie}{i.\,e.,\xspace}

\renewcommand{\vec}[1]{\bm{#1}}
\DeclareMathOperator*{\argmin}{argmin}
\crefname{section}{Sec.}{Sections}
\crefname{figure}{Fig.}{Figure}
\crefname{table}{Tab.}{Table}
\crefname{equation}{Equ.}{Equation}

\definecolor{faublue}{RGB}{0,51,102}

\newcommand{\bx}{\vec{x}}

\newcommand{\map}{mAP\xspace}

\newcommand{\icdars}{\textsc{Icdar17-WI}\xspace}

\newcommand{\clamm}{CLaMM\xspace}
\newcommand{\clammsix}{\textsc{Clamm16}\xspace}
\newcommand{\clammseven}{\textsc{Clamm17}\xspace}

\definecolor{faublue}{RGB}{0,51,102}
\definecolor{darkgreen}{rgb}{0,0.6,0}
\definecolor{bblue}{HTML}{4F81BD}
\definecolor{rred}{HTML}{C0504D}
\definecolor{ggreen}{HTML}{9BBB59}
\definecolor{ppurple}{HTML}{9F4C7C}

\NewDocumentCommand{\rot}{O{60} O{1em} m}{\makebox[#2][l]{\rotatebox{#1}{#3}}}%
\NewDocumentCommand{\rotn}{O{90} O{1em} m}{\makebox[#2][l]{\rotatebox{#1}{#3}}}%
\NewDocumentCommand{\rotninety}{O{90} O{1em} m}{\makebox[#2][l]{\rotatebox{#1}{#3}}}%

\newcommand{\myerrorplot}[2]{%
\addplot+[name path=upper, thick, draw=none] table
[x expr=(\coordindex+1), y expr={(100-(\thisrowno{0} + \thisrowno{1})*100}, col sep=comma]
{%
	#1%
};
\addplot+[name path=lower, thick, draw=none] table
[x expr=(\coordindex+1), y expr={(100-(\thisrowno{0} - \thisrowno{1})*100}, col sep=comma]
{%
	#1%
};
\addplot+[fill=#2, fill opacity=0.3, draw opacity=0] fill between[of=upper and lower];
}

\begin{document}
\title{Deep Generalized Max Pooling}

\author{\IEEEauthorblockN{Vincent Christlein\IEEEauthorrefmark{1}, Lukas Spranger\IEEEauthorrefmark{1}, 
Mathias Seuret\IEEEauthorrefmark{1}, 
Anguelos Nicolaou\IEEEauthorrefmark{1},
Pavel Král\IEEEauthorrefmark{2}, Andreas Maier\IEEEauthorrefmark{1}}
\IEEEauthorblockA{
\IEEEauthorrefmark{1}Pattern Recognition Lab, Friedrich-Alexander-Universität Erlangen-Nürnberg,
91058 Erlangen, Germany\\
\{firstname.lastname\}@fau.de
}
\IEEEauthorblockA{
\IEEEauthorrefmark{2} Dept of Computer Science \& Engineering, University of West Bohemia
Plzeň, Czech Republic\\
pkral@kiv.zcu.cz
}%
\thanks{This work has been partly supported by the Cross-border Cooperation Program Czech Republic -- Free State of Bavaria ETS Objective 2014-2020 (project no. 211) and project LO1506 of the Czech Ministry of Education, Youth and Sports.}%
}

\maketitle

\begin{abstract}
Global pooling layers are an essential part of Convolutional Neural Networks
(CNN). They are used to aggregate activations of spatial locations to produce a
fixed-size vector in several state-of-the-art CNNs. 
Global average pooling or global max pooling are commonly used for converting convolutional features of variable size images to a fix-sized embedding. 
However, both pooling layer types are computed spatially independent: each individual activation map is pooled and thus activations of different locations are pooled together. 
In contrast, we propose Deep Generalized Max Pooling that
balances the contribution of all activations of a spatially coherent region
by re-weighting all descriptors so that the impact of frequent and rare ones is equalized. 
We show that this layer is superior to both average and max pooling on the classification of Latin medieval manuscripts (CLAMM'16, CLAMM'17), as well as writer identification (Historical-WI'17).
\end{abstract}

\begin{IEEEkeywords}
pooling; deep learning; document analysis, document image classification, writer identification
\end{IEEEkeywords}

\IEEEpeerreviewmaketitle

\section{Introduction}

Convolutional neural networks (CNN) have demonstrated remarkable performance in many computer vision tasks in the past years, and are also predominant in the field of document analysis, such as font recognition~\cite{Tensmeyer17}, word spotting~\cite{Sudholt16} or writer identification~\cite{Christlein17ICDAR}.
Many successful CNN architectures compute a global image descriptor at the final stage of their computational graph by aggregating the feature maps using global average pooling~\cite{He16a,Huang17} or global max pooling~\cite{Oquab2015a,Sudholt16}. These layers also allow the use of images of arbitrary dimensions.

Conceptually, one has to differentiate between %
average/max pooling used for downsampling that pools over local descriptors extracted from different image regions, and global average/max pooling used in CNNs that pool the activations over the entire activation map. 
Commonly, each activation map is pooled independently. It follows for global pooling that the final representation contains activations from different locations. 
For highly structured images, global average pooling might give too much emphasis on frequently occurring patches in the input image, \eg the document background or the script. While global max pooling does not have this problem, it might get influenced by noise.

\begin{figure}[t]
    \centering
    \bigskip
    \includegraphics[width=0.8\columnwidth]{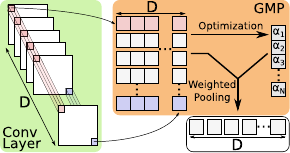}
    \caption{Overview of Deep Generalized Max Pooling. The activation volume that is computed from a convolutional layer serves as input for the DGMP layer. A linear optimization problem with D unknowns is solved using each local activation vector along the depth axis of the activation volume as linear equation with D unknowns. The output is a weighted sum of the local activation vectors.}
    \label{fig:gmp_overview}
\end{figure}
We propose a new global pooling that aggregates activations in a spatially coherent way.
Therefore, a weighted aggregation of local activation vectors is applied, where a local activation vector is defined to be a position in the activation volume along the depth dimension, see \cref{fig:gmp_overview} (left). In contrast to global average/max pooling, activations of specific locations are balanced jointly with each other.

The weights for each local vector are computed through an optimization process, where we chose the use
of Generalized Max Pooling (GMP)~\cite{Murray16}. It was originally developed to balance local descriptors in a traditional Bag-of-Words (BoW) setting, where a global representation is computed from embeddings of local descriptors. 
Average pooling is commonly used to aggregate local embeddings. Yet, average pooling may suffer from over-frequent local descriptors computed from very similar or identical image regions. 
GMP balances the influence of frequent and rare descriptors according to the global descriptor similarity. 
We propose the use of GMP in a deep learning setting, and phrase GMP as a network layer, denoted as Deep Generalized Max Pooling (DGMP).

This can also be regarded as a form of statistical attention. While attention depends on the content, we balance activation locations independent of the content but enforcing that the image-to-image similarity becomes independent of frequent and rare image locations. We believe that a combination with attention could further improve the accuracy. 

In detail, our contributions are as follows:
\begin{enumerate*}
    \item We propose a spatially coherent pooling by weighting local activation vectors. For the computation of the weights, we propose to use DGMP. It balances frequent and rare embeddings that consist of locally coherent activations. The code for the DGMP layer is publicly available.\footnote{\url{https://github.com/VChristlein/dgmp}}
    \item Furthermore, the proposed DGMP layer is evaluated on two different scenarios: 
    \begin{enumerate*}
        \item[(i)] writer identification/retrieval and 
        \item[(ii)] script type classification
    \end{enumerate*}. 
    For writer identification, we directly use the output of the DGMP layer to compute a similarity while we add a fully-connected layer on top of the DGMP layer for the classification of cursive Latin medieval fonts. 
    \item For both scenarios, we show that DGMP is superior to other global pooling methods, such as global average or max pooling. 
    \item While not being the focus of the paper, our results achieve better or comparable results to state-of-the-art methods for font classification.
\end{enumerate*}

The paper is organized as follows. After giving an overview of the related work
in pooling techniques and historical document classification in \cref{sec:related_work}, we explain the methodology and how we phrase GMP as a network layer in \cref{sec:methodology}. In \cref{sec:evaluation}, the proposed pooling layer is compared against global average and global max pooling using different scenarios and datasets. We show that DGMP is a powerful replacement for both of them by adding only one single additional parameter to the network.

\section{Related Work}
\label{sec:related_work}
\subsection{Pooling}
In the traditional Bag-of-(visual)-Words (BoW) model, local descriptors are aggregated to form a global image descriptor. Commonly, this is achieved by using average pooling. However, average pooling assumes that the descriptors in an image are independent and identically distributed. Thus, more frequently occurring descriptors will be more influential in the final representation and dominate the similarity. 
This phenomenon is known as visual burstiness~\cite{Jegou09}. Bursty descriptors can dominate the similarity metric while being irrelevant to distinguish different samples. For example in document images, the image background, \ie areas without script, commonly do not contribute to the class decision. 
A common technique to counter burstiness is to normalize the final global descriptor, \eg by \emph{power normalization}~\cite{Perronnin10IFK}. Exceptions of this are the works of \etal{Murray}~\cite{Murray14} and \etal{Jégou}~\cite{Jegou14TEA} who proposed \emph{Generalized Max pooling} and \emph{Democratic Aggregation}, respectively. Both methods re-weight the descriptors such that their influence regarding the final similarity score is equalized. When comparing both methods, \etal{Murray}~\cite{Murray16} found that GMP is superior to democratic aggregation. 

In neural networks, we have to differentiate between \emph{spatial} pooling and \emph{global} pooling. Spatial pooling fuses information of input across spatial
locations. Thereby the number of parameters is decreased and thus the computational cost. It  also provides some affine invariance and mitigates the risk of overfitting.
Spatial max pooling was already used in the seminal work of \etal{LeCun}~\cite{LeCun98} and was a typical building block of CNNs until recently. Nowadays, it is more common to increase the stride of the convolution.

The use of a global average layer as a last layer was proposed by \etal{Lin}~\cite{lin2013network}, and got its breakthrough by the well-known residual network (ResNet)~\cite{He16a}. Global max pooling was for example proposed for weakly-supervised learning~\cite{Oquab14} and is also used in the PHOCNet for the task of word spotting~\cite{Sudholt16}. 

It is also possible to learn a pooling function from the data using the neural network optimization framework. \etal{Lee}~\cite{Lee16} suggest a tree-based spatial pooling, where pooling filters are learned that are structured as a tree. The authors compare the method also with a mixed pooling layer, a simple weighted sum of average and max pooling, \ie:
\begin{equation}
\bm{\xi}_\text{mix} = \alpha \cdot \bm{\xi}_\text{max} + (1-\alpha) \cdot \bm{\xi}_\text{avg}   \label{eq:mixed}
\end{equation}
where $\alpha$ is learned by backpropagation and $\bm{\xi}_\text{max}$, $\bm{\xi}_\text{avg}$ the max and average pooling representations.
Another pooling method~\cite{Pinheiro15}, denoted as log-sum-exp (LSE) pooling, computes a smooth approximation of the max function. It is computed for each feature map $\bm{\Phi}_i$ as follows:
\begin{equation}
\bm{\xi}_\text{lse,i} = \frac{1}{r} \cdot \log \left[ \frac{1}{\lvert\bm{\Phi}_i\rvert} \sum_{x \in \bm{\Phi}_i} \exp(r\,x)\right] \;,
\label{eq:lse}
\end{equation}
where $r$ is a pooling parameter (commonly not learned during training) that allows a transition from average pooling for $r \to 0$ to max pooling for $r \to \infty$. 
All these methods work on the basis of a single individual activation map without notion of the other activation maps. In contrast, we propose DGMP, a network layer that balances frequent and infrequent local activation vectors along the depth dimension.

\subsection{Historical Document Image Classification}

While we propose a general pooling layer, which can be used in most common network architectures     by replacing the global average pooling layer, we believe that the classification of writings benefits the most of our DGMP layer due to the repetitive character of script. Balancing frequent activations from the background with script features that are used to classify the document can be helpful to improve the image embedding.

We mainly focus on two specific tasks: Script type or font classification and writer identification/retrieval. Script type classification is an important aspect of paleographic research~\cite{Stutzmann16, Kestemont17}, and has recently gained more attention through the ICFHR'16 and ICDAR'17 competitions in the classification of Latin medieval manuscripts (\clamm)~\cite{Cloppet16,Cloppet17}. 
\etal{Tensmeyer}~\cite{Tensmeyer17} proposed the use of an ensemble of CNNs trained with different network architectures (for \clammsix) and image scales (for \clammseven). They employ a patch-based approach for training as well as testing. They show that they could improve the recognition accuracy by means of individual foreground/background brightness augmentation. Another patch-based approach achieves similar results by standardizing the images using zero component analysis~\cite{Christlein18PHD}. In contrast, we show that we can achieve superior results for the \clammsix dataset and comparable results for the \clammseven dataset by single image views. 

Another line of research is writer identification, where the task is to find the correct writer of a query sample given a dataset of samples with known scribes~\cite{Christlein17ICDAR}. The evaluation procedure for writer retrieval is similar, however, the full ranking of the dataset is evaluated in terms of mean average precision (\map). 
The most deep learning approaches compute CNN activation features for local image patches 
using the penultimate layer of a CNN using the writers of the training set as targets~\cite{Fiel15CAIP,Christlein15GCPR,Christlein17ICDAR,Tang16,Chen19}. These local feature descriptors are subsequently aggregated to form a global embedding and then used for comparison.  A semi-supervised learning scheme is suggested by \etal{Chen}~\cite{Chen19} which makes use of additional unlabeled data. In comparison, \etal{Christlein}~\cite{Christlein17ICDAR} proposed to use an unsupervised learning scheme to compute deep activation features that are eventually encoded using VLAD~\cite{Jegou12ALI}. In a subsequent work~\cite{Christlein18DAS}, they show that GMP improves the encoding consistently. \etal{He}~\cite{He19} employ auxiliary tasks to improve writer identification of single word images.
Direct embeddings computed for example via triplet networks~\cite{Keglevic18,Cloppet17} are also used. In these cases, no additional feature encoding such as VLAD is needed. We also rely on a triplet network, \ie we use a triplet loss to learn an image embedding that can directly be used for comparison. 

\section{Methodology}
\label{sec:methodology}
Our network layer is based upon \emph{generalized max pooling}
(GMP)~\cite{Murray16}. 
It is a pooling technique that generalizes sum pooling and max pooling. The goal
is to compute a single vector representation, from a set of local embeddings.
Generalized max pooling was proposed as an aggregation function that summarizes
feature embeddings obtained by embedding functions. The embedding function can
be VLAD, Fisher vectors or CNN codes. However, GMP was originally not proposed
as a neural network layer. We reinterpret generalized max pooling as a global
pooling layer and incorporate our GMP layer in neural network architectures --
replacing the commonly used global average pooling. Other than the original
formulation, this allows an end-to-end training using backpropagation.

\subsection{Generalized Max Pooling}
The standard approach for computing a global descriptor of an image given a set of
local descriptors consists of two steps: an \emph{embedding} step and an
\emph{aggregation} phase. In the embedding phase, an
embedding function $\phi$ maps each local descriptor into a high-dimensional
space. During the aggregation step an aggregation function $\psi$, \eg
sum pooling or max pooling, computes a single vector from the embedded feature
vectors.

If we use sum pooling, the global descriptor for a set of local descriptors
$\mathcal{X}=\{\bx_i \in \mathbb{R}^D, i=1,\ldots,N\}$ becomes 
\begin{equation}
\bm{\xi}_\text{sum} = \psi(\phi(\mathcal{X})) = \sum_{\bm{x} \in \mathcal{X}} \phi(\bm{x})\;.
\end{equation}
Since we sum over all descriptors, the aggregated descriptors can
suffer from interference of unrelated descriptors that influence the similarity,
even if they have low individual similarity. Assume the dot-product is used to
compute the similarity between two images that are represented by the sets
$\mathcal{X}$ and $\mathcal{Y}$, respectively. Their similarity
$\mathcal{K}(\mathcal{X}, \mathcal{Y})$ follows as:
\begin{equation} \mathcal{K}(\mathcal{X}, \mathcal{Y}) = \sum_{\bm{x} \in
	\mathcal{X}} \sum_{\bm{y} \in \mathcal{Y}} \phi(\bm{x})^T
	\phi(\bm{y})\;.
\end{equation}
Hence, the similarity is defined by the sum of pairwise similarities between all
descriptors, \ie unrelated descriptors with low similarity contribute to the
overall expression.

\etal{Murray}\cite{Murray16} propose to perform a weighted sum
pooling. For every $\bm{x} \in \mathcal{X}$ a weight $\alpha
(\bm{x})$ is introduced. The weights shall equalize the contribution of each
element to the similarity scores. The aggregated representation
$\bm{\xi}$ becomes:
\begin{equation}
\bm{\xi}_\text{gmp} = \sum_{\bm{x} \in \mathcal{X}} \alpha(\bm{x}) \phi(\bm{x})
= \bm{\Phi} \, \bm{\alpha} \;,
\label{eq:gmp_weighted_sum}
\end{equation}
where $\bm{\Phi} \in \mathbb{R}^{D\times N}$ and $\bm{\alpha}\in\mathbb{R}^{N\times1}$.

Generalized max pooling is a procedure to obtain the weights
$\bm{\alpha}$ by solving an optimization problem. 
The goal is to enforce the dot-product similarity between the aggregated
representation $\bm{\xi}_{gmp}$ and each patch encododing
$\phi$ to be a constant value. The value of the constant
(we used $1$) has no influence, since the final encoding will be normalized.
Thus, we end up with a linear system of $N$ 
equations and $D$ unknowns.
\begin{equation}
	\bm{\Phi(x)}^\top \bm{\xi}_{\text{gmp}} = \bm{1}_N \,,
\end{equation}
where $\vec{1}_N$ denotes the vector of all constants set to $1$. 
This linear system can be turned into a least-squares ridge regression problem:
\begin{equation}
	\bm{\xi}_{\text{gmp}} = \argmin_{\bm{\xi}} 
	\lVert \bm{\Phi}^\top\bm{\xi} - \bm{1}_N \rVert_2^2 +
	\lambda\lVert\bm{\xi}\rVert_2^2\;,
	\label{eq:primal}
\end{equation}
with $\lambda$ being a regularization term that stabilizes the solution. An
interesting property of $\lambda$ is that the pooling resembles sum-pooling when $\lambda\to\infty$. In case of $\lambda\to 0$, we turn the ridge regression to a least-squares regression problem.

\subsection{Deep GMP Layer}
We adopt generalized max pooling as a trainable neural network layer following the known-operator paradigm \cite{maier2018-precision}. 
This allows for an end-to-end training given images and a loss function, such as softmax or triplet loss. 
Both generalized max pooling and average pooling are at their core variants of
weighted sum pooling. Therefore, we can easily interpret GMP as a form of global
pooling, similar to global average pooling.  Assume the final activation volume
of a convolutional neural network with height $h$, width $w$ and depth $d$
(= number of activation maps), see \cref{fig:gmp_overview}. 
We interpret the vectors consisting of the
activations at a location $i$ with $i=\{1,\ldots,h\cdot w\}$
across the whole depth $d$ as the local descriptor
embedding $\bm{\phi}_{i}$. That means, there are $N = h \cdot w$
embeddings $\bm{\phi}$ of dimension $D=d$ that are aggregated. 

In other words, we compute a weighted sum over all locations in the
activation volume, yielding a $D$-dimensional vector -- the global descriptor
$\bm{\xi}$. %
This is substantially different from average or max pooling that average or max-pool among one activation map without considering the other activation maps.

\begin{figure}[t]
\centering
		\begin{tikzpicture}
			\begin{scope}[spy using outlines={magnification=3, height=1.4cm, width=2.2cm},
				connect spies,
				every spy in node/.style={
					rectangle,
					draw, 
				ultra thick, cap=round}]
				\node[inner sep=0](i1){
					\includegraphics[height=.12\textheight]{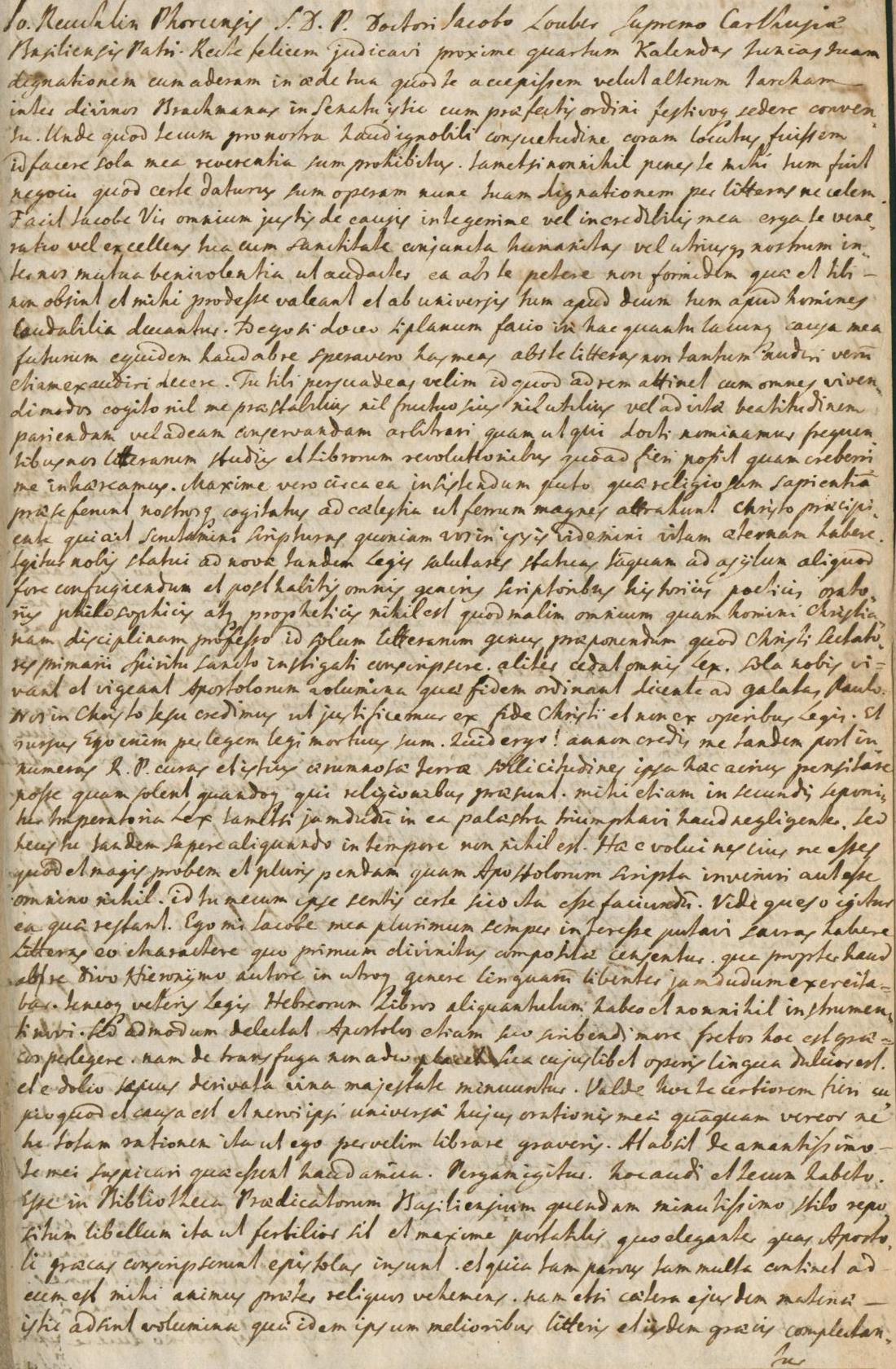} %
				};
				\node[right=3.2cm of i1,inner sep=0](i2){
					\includegraphics[height=.12\textheight]{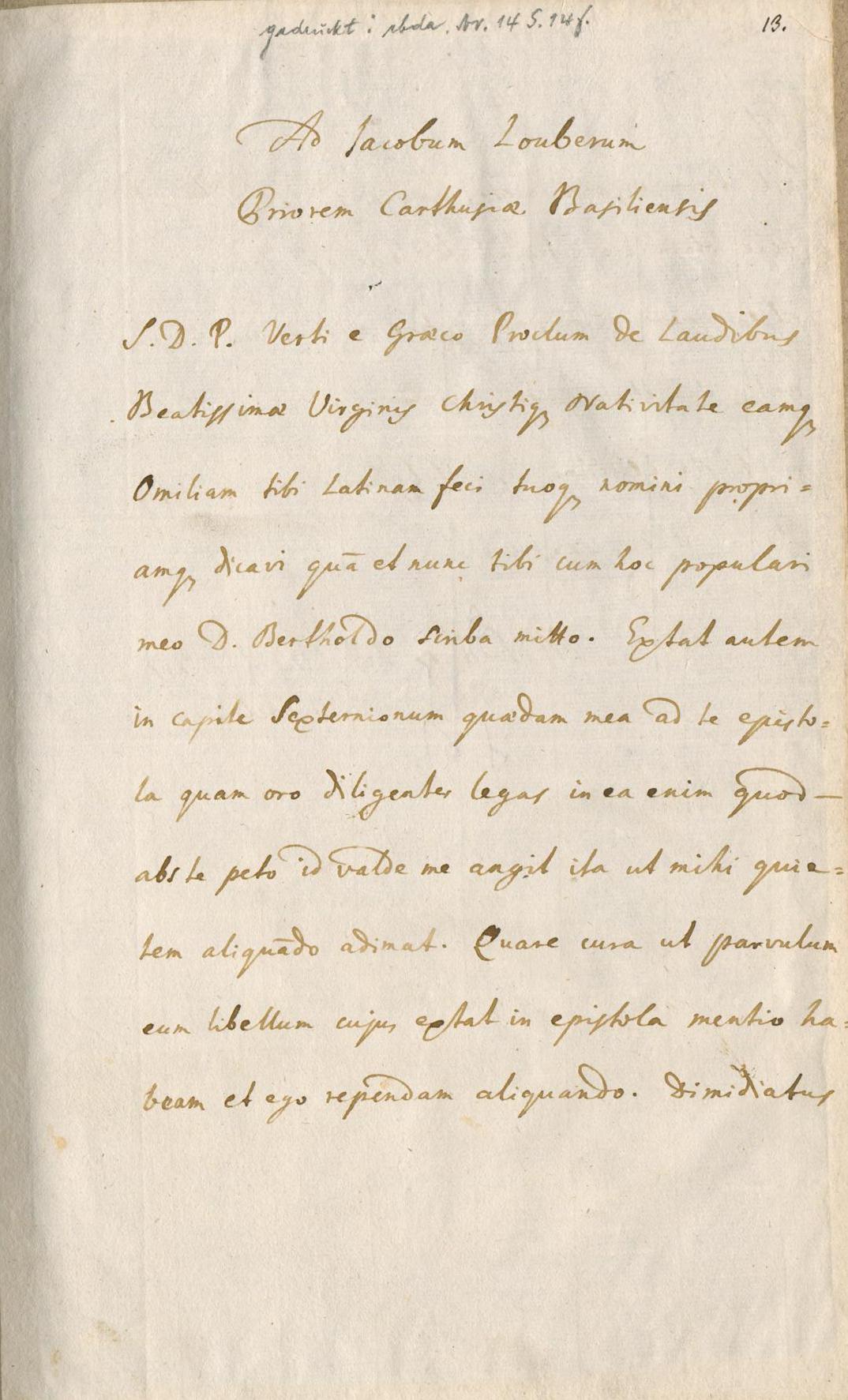} 
				};
				\node[below=0.7cm of i1.south east,anchor=north east,inner sep=0](i3){
					\includegraphics[height=.12\textheight]{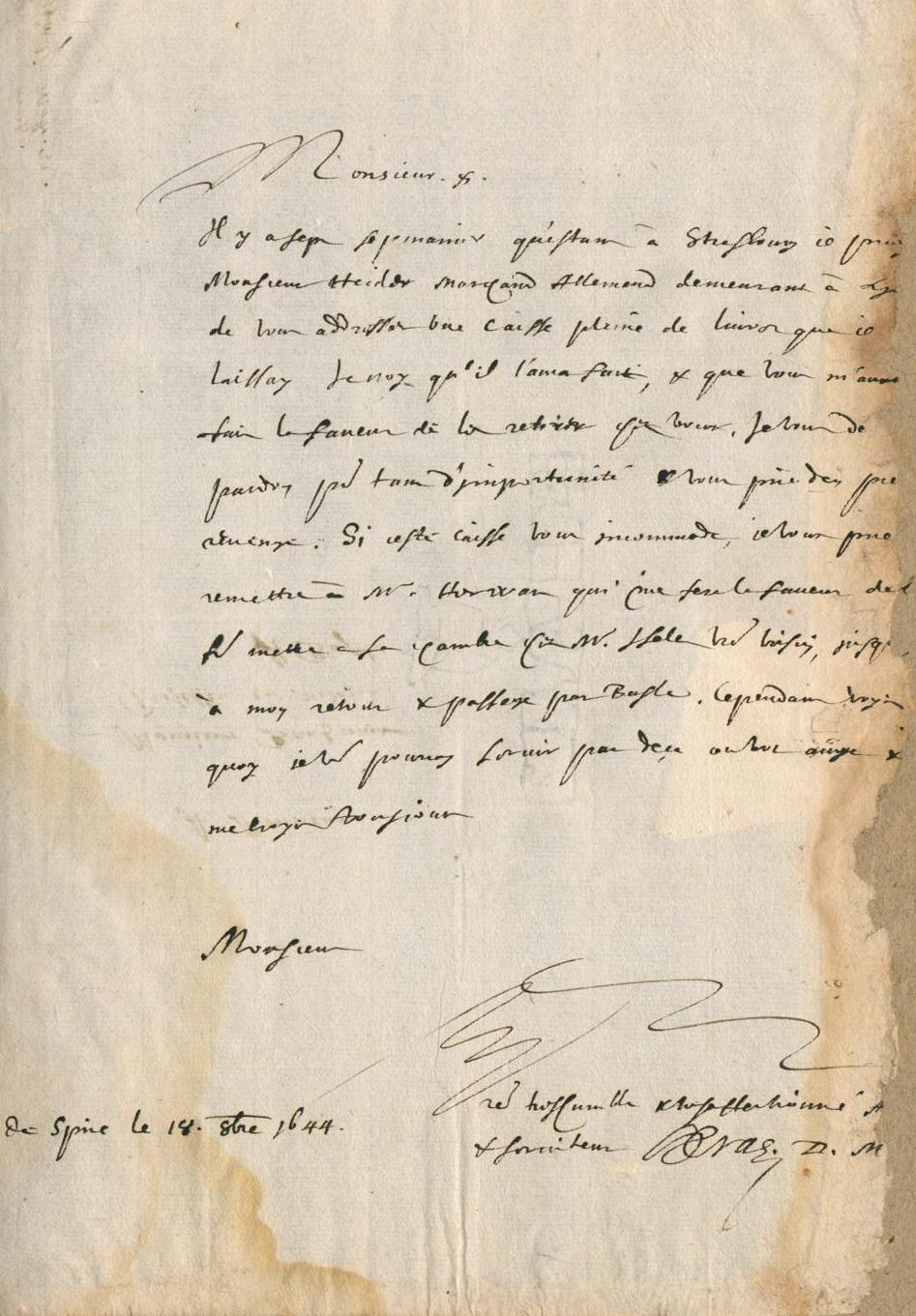}%
				};
				\node[below=0.7cm of i2.south west,anchor=north west,inner sep=0](i4){
					\includegraphics[height=.12\textheight]{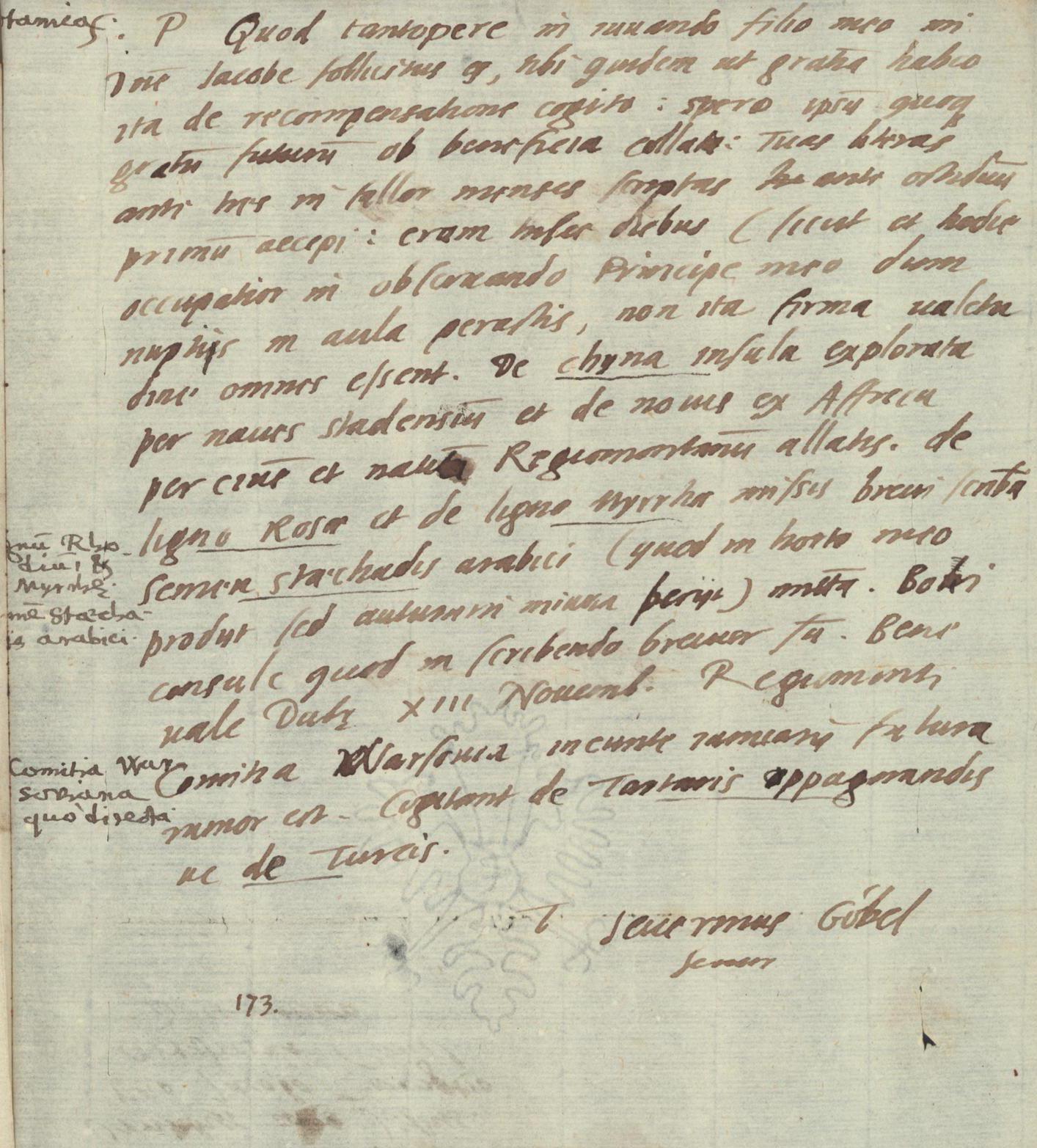} 
				};
				\begin{scope}[x={(i1.south east)},y={(i1.north west)}]
					\spy[blue] on (-540/1680, -800/1100) 
					in node[anchor=south,yshift=-0.2cm] at ($(i1.south east)!0.5!(i2.south west)$); 
				\end{scope}
				\begin{scope}[x={(i2.south east)},y={(i2.north west)}] 
					\spy[blue] on ($(i2) + (-300/1680,400/1100)$)
					in node[anchor=north] at ($(i1.north east)!0.5!(i2.north west)$); 
				\end{scope}
				\begin{scope}[x={(i3.south east)},y={(i3.north west)}]
					\spy[orange] on (i3) %
					in node[anchor=south] at ($(i3.south east)!0.5!(i4.south west)$); 
				\end{scope}
				\begin{scope}[x={(i4.south east)},y={(i4.north west)}]
					\spy[magenta] on ($(i4) + (-400/1680,400/1100)$)
					in node[anchor=north,yshift=0.2cm] at ($(i3.north east)!0.5!(i4.north west)$); 
				\end{scope}
			\end{scope}
				\node[below=0cm of i1]{(a)\phantomsubcaption\label{fig:w11}};
				\node[below=0cm of i2]{(b)\phantomsubcaption\label{fig:w12}};
				\node[below=0cm of i3]{(c)\phantomsubcaption\label{fig:w2}};
				\node[below=0cm of i4]{(d)\phantomsubcaption\label{fig:w3}};
        \end{tikzpicture}
		\caption{\icdars indicative samples 
			Samples \subref{fig:w11} and \subref{fig:w12} stem from
			the same writer (acc.\ to the ground truth) while samples \subref{fig:w2} and \subref{fig:w3} come from two different ones (IDs: 11-3-IMG\_MAX\_1005484, 11-3-IMG\_MAX\_1005478, 358-3-IMG\_MAX\_1031951, 7-3-IMG\_MAX\_10051).}
		\label{fig:icdar_wi_samples}
\end{figure}
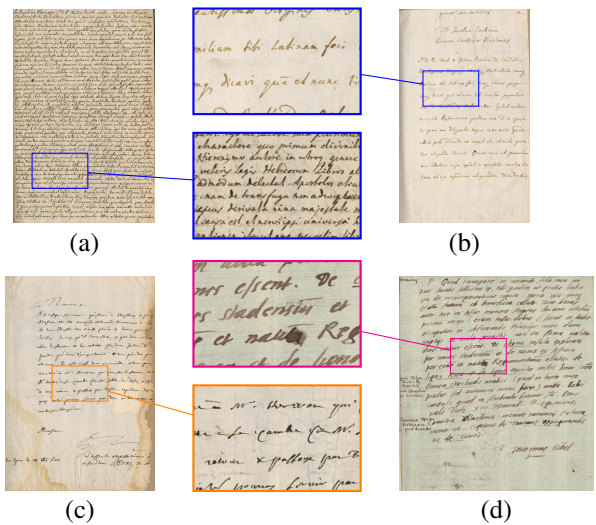

We follow \etal{Murray}~\cite{Murray16} and solve the dual formulation for each sample to explicitly compute the weights. Let $\bm{\xi}=\vec{\Phi}\vec{\alpha}$ and insert it into \cref{eq:primal}, it follows:
\begin{align}
\begin{split}
\bm{\alpha}_{gmp, \lambda} = {}& \argmin_{\bm{\alpha}} \lVert \bm{\Phi}^T\bm{\Phi} \bm{\alpha} - \bm{1}_n \rVert^2 + \lambda \lVert \bm{\Phi}\bm{\alpha} \rVert^2
\\
={}& 
\argmin_{\bm{\alpha}} \lVert \bm{K} \bm{\alpha} - \bm{1}_n \rVert^2 + \lambda
\bm{\alpha}^\top \bm{K} \bm{\alpha} \;,
\end{split}
\end{align}
where $\bm{K}$ is the Gram matrix, yielding the closed-form solution:
\begin{align}
\bm{\alpha} = (\bm{K} + \lambda \bm{I}_n)^{-1} \bm{1}_n \;,
\label{eq-gmp-dual}
\end{align}
which does not rely on the embeddings but on a similarity kernel. The GMP representation follows by \cref{eq:gmp_weighted_sum}.
Note that for $N \gg D$ the solution of the primal (where also a closed-form solution exists) is more efficient. In our case $D < N$ for the normal ResNet50 version with $400\times 400$ input image sizes and $N \approx D$ for the modified ResNet50 version that we used for script type classification.
In the original formulation, $\lambda$ is a hyperparameter that has to be
cross-validated. Since all operations involved are differentiable, we
can optimize for $\lambda$. Yet, the right initialization of $\lambda$ still matters to some degree, see our experiments.
The output of the DGMP layer is eventually normalized such that its
$\ell^2$ norm equals to one.

\section{Evaluation}
\label{sec:evaluation}
We evaluate DGMP on two different tasks: 
\begin{enumerate*}
    \item writer identification/retrieval and \item script type classification.
\end{enumerate*}
For both scenarios, we show that DGMP is superior to other
pooling mechanisms, such as average or max pooling. 

\subsection{Datasets}
We evaluate the following publicly available datasets:
\subsubsection{\icdars} is a dataset composed of letters used in the ICDAR 2017 Writer Identification competition~\cite{Fiel17ICDAR}.
The training set contains 394 writers contributing three samples each. 720 other writers with five samples each are provided for testing.
In \cref{fig:icdar_wi_samples}, indicative examples of the \icdars dataset can be seen.
The task consists of training a representation given the writer-disjoint training set and use that representation to effectively retrieve the test-set.
It can be seen that writer identity is hard to infer as general look and feel of the document dominates the visual similarity.
\begin{figure}[t]
    \centering    
    \begin{tikzpicture}
			\begin{scope}[spy using outlines={magnification=3, height=1.4cm, width=2.2cm},
				connect spies,
				every spy in node/.style={
					rectangle,
					draw, 
				ultra thick, cap=round}]
				\node[inner sep=0](i1){
				\includegraphics[height=.12\textheight]{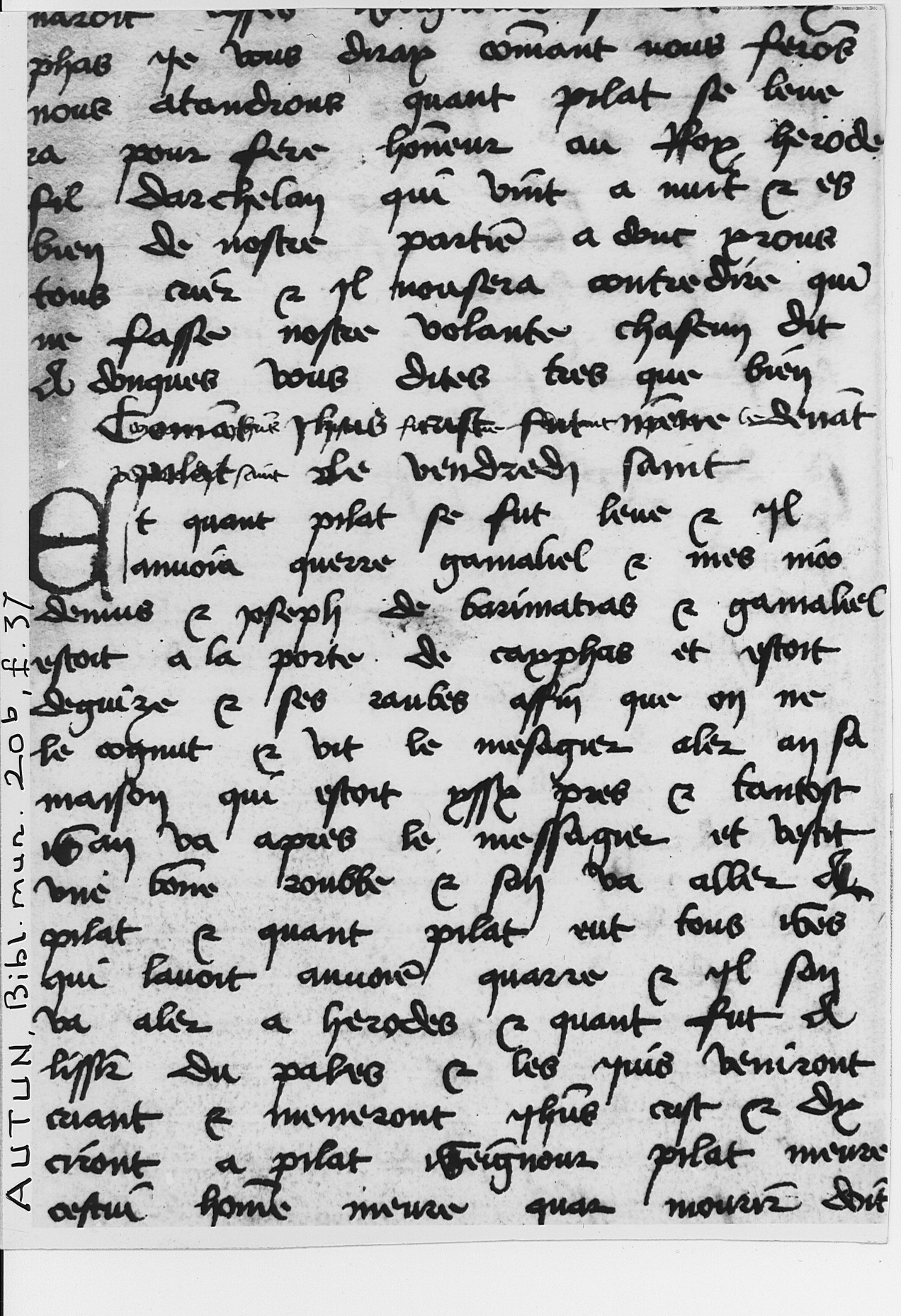}
				};
				\node[right=3.2cm of i1,inner sep=0](i2){
					\includegraphics[height=.12\textheight]{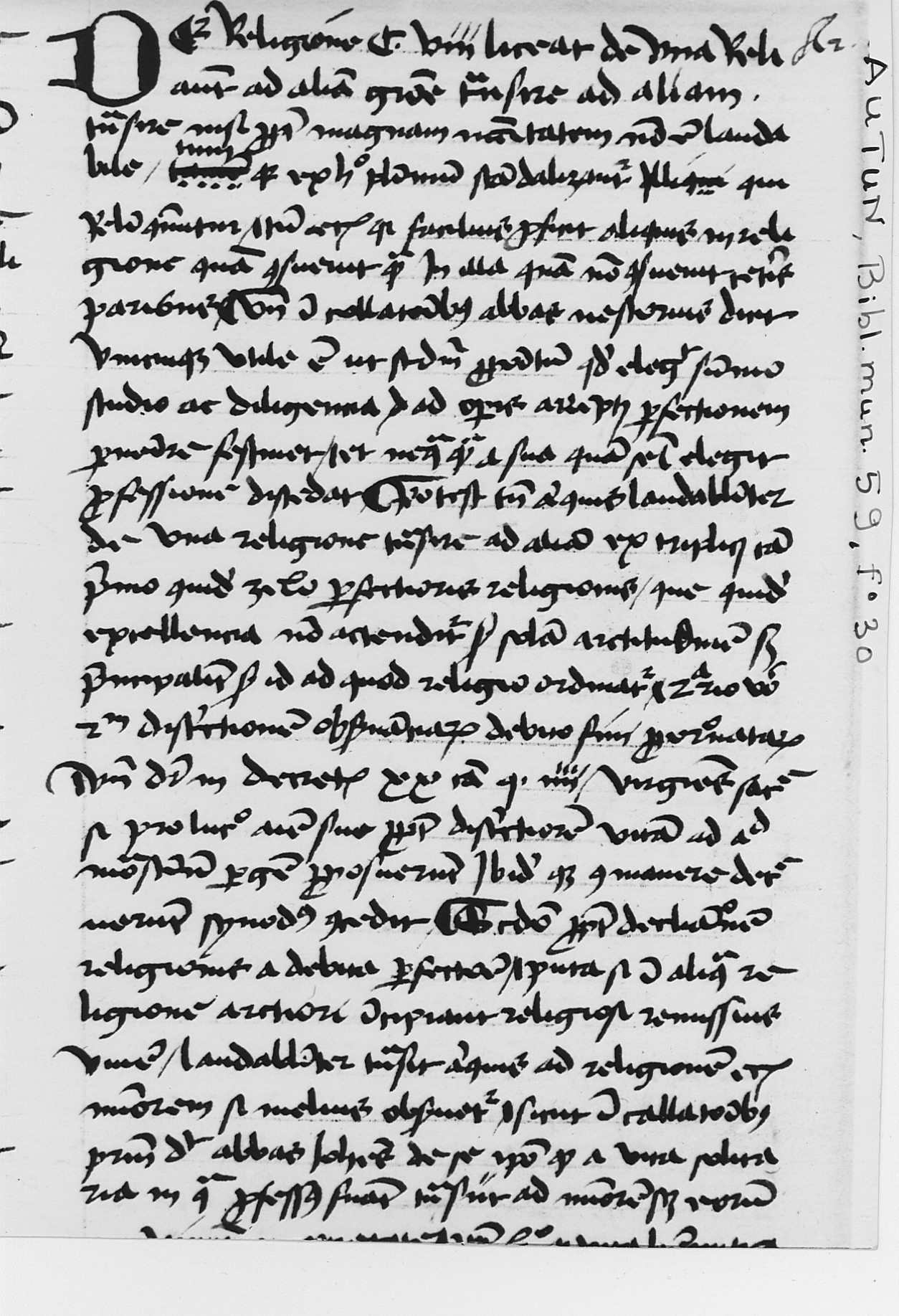}	
				};
    			\begin{scope}[x={(i1.south east)},y={(i1.north west)}]
					\spy[orange] on (i1) %
					in node[anchor=south,yshift=-0.05cm] at ($(i1.south east)!0.5!(i2.south west)$); 
				\end{scope}
				\begin{scope}[x={(i2.south east)},y={(i2.north west)}]
					\spy[magenta] on ($(i2) + (-400/1680,400/1100)$)
					in node[anchor=north,yshift=0.05cm] at ($(i1.north east)!0.5!(i2.north west)$); 
				\end{scope}
			\end{scope}
				\node[below=0cm of i1]{(a) Cursiva\phantomsubcaption\label{fig:cl1}};
				\node[below=0cm of i2]{(b) Semihybrida\phantomsubcaption\label{fig:cl2}};
        \end{tikzpicture}
        \caption{\clammsix samples of two similar fonts (excerpts of IRHT\_P\_000012, IRHT\_P\_000020).}
    \label{fig:clamm_samples}
\end{figure}
\subsubsection{\clammsix and \clammseven} were introduced in the competitions on Classification of Medieval Latin Manuscripts (\clamm) of 2016~\cite{Cloppet16} and 2017~\cite{Cloppet17}, respectively. 
The task is image classification and both datasets contain twelve classes representing script types. 
The \clammsix dataset consists of 2000 training and 1000 test samples while the \clammseven dataset contains 3540 and 2000 samples, respectively. Example images can be seen in \cref{fig:clamm_samples}.

\subsection{Error Metrics} 
We report top-1 accuracy for writer identification and font classification. For writer retrieval, the mean average precision (\map) is given. The latter is computed by taking the mean of computing the average precision per query sample. The average precision for a specific query sample $q$ and $R$ relevant samples in a dataset of size $S$ is given as: 
\begin{equation}
  \mathrm{AP}_q = \frac{1}{R} \sum_{r=1}^S \mathrm{Pr}_q(r) \cdot
  \mathrm{rel}_q(r) \;,
\end{equation}
where $\mathrm{rel}_q(r)$ is an indicator function returning $1$ if the document
at rank $r$ is relevant, and $0$ otherwise, and $\mathrm{Pr}_q(r)$ is the 
precision at rank $r$. 

\subsection{Experiments}
For all our experiments, we use the ResNet-50~\cite{He16a} architecture with weights
pretrained on ImageNet (provided by PyTorch). 
The last fully connected layer is either changed to
represent the twelve classes of the \clamm datasets or omitted in case of writer
identification. The global pooling layer (= penultimate or last layer) is set to
the respective pooling layer to be evaluated. 
We first evaluate GMP for writer identification, and then for script type
classification. 

\subsubsection{Writer Identification}
For writer identification, we are interested in global descriptors that discriminate different writers, \ie we use directly the pooled representation as our image descriptor. However, exploratory experiments on a validation split of the \icdars training dataset using the full, colored images did not achieve good results. 
Since the images in the \icdars dataset are rather large, and we want to distinguish the authors on the individual characteristics of their handwriting, which are typically only present in small image portions, the image is subdivided into patches of size $400 \times 400$ with a stride of 256.
Crops that do not contain any handwriting are filtered out by setting a threshold (=~2000) on the sum of the Canny edge detector~\cite{Canny} output.
We use the binarized version of the images to be independent of pen and background variations, and do not employ any additional transformations or augmentation techniques apart from standardizing the data by enforcing a zero-mean and standard deviation of one.

During test time, images are also sub-divided into overlapping patches and standardized.
The respective descriptors computed by our model are then averaged to obtain the overall image descriptor for computing the ranking based on the query images.

The networks are trained by means of the triplet loss~\cite{Schroff15}: 
\begin{equation}
L_{\text{tri}} = \sum_{ \substack {a, p, n \\ 
                       y_a = y_p, y_a \neq y_n}} 
                       [m + D(\phi(a), \phi(p)) - D(\phi(a), \phi(n))]_{+} \;, 
\end{equation}
where $[\cdot]_{+} = \max(0, \cdot)$ and $D(x,y)$ denotes a distance metric between two samples $x$ and $y$. The triplet loss enforces that for a given
anchor $a$, the embedding of a positive sample $p$ that belongs to the same class
$y_p = y_a$ is closer to the encoding of the anchor than a negative sample $n$
belonging to a different class $y_n$ by at least a margin $m$.

We use the hard-batch variant of the triplet loss~\cite{Hermans17}, \ie the hardest positive and hardest negative pairs per (mini-)batch are selected for the triplet creation. 
Therefore, we construct the (mini-)batches of size $P \cdot K$ by randomly sampling $P$ writers and $K$ image crops for each writer. In all our experiments on the historical writer identification dataset, we set $P = 14$ and $K = 4$. We evaluate the impact of different margins for the triplet loss on the performance of our model. 

For optimization, we use the AMSGrad variant~\cite{Reddi18} of
Adam~\cite{Kingma15} with default parameters ($\beta_1 = 0.9, \beta_2 = 0.999$) and a weight decay of $10^{-5}$. 
We employ a learning rate of $\eta = 2 \cdot 10^{-4}$. We recognized a slight improvement when using a higher learning rate for the GMP pooling parameter (learning rate is multiplied by $10^3$).
We run the experiments for 300 epochs and employ an exponential learning rate decay starting at epoch 150 as suggested by \etal{Hermans}~\cite{Hermans17}. 
As validation set, we choose 40 writers ($\approx$\SI{10}{\percent}) of the official training set. The best model is selected according to the best validation error along the 300 epochs.

\begin{table}[t]
\centering
\caption{Impact of different triplet loss margins evaluated on the \icdars dataset for global average pooling (Avg), global max pooling (Max), and deep generalized max pooling (DGMP) with different initializations of $\lambda$.}
\begin{tabular}{l*{6}c}
\toprule
  & \multicolumn{2}{c}{margin 0.1} &  \multicolumn{2}{c}{margin 0.2} & \multicolumn{2}{c}{margin 0.5}\\
  \cmidrule(l){2-3} \cmidrule(l){4-5} \cmidrule(l){6-7}
    & \map   & top-1 &  \map  &  top-1  & \map & top-1\\ \midrule
Avg         & 50.44 & 69.19  & 48.75 & 67.61  & 48.60 & 67.19 \\
Max         & 50.32 & 69.63  & 46.63 & 65.47  & 40.04 & 57.03 \\
DGMP, $\lambda=1$  & 53.76 & 72.19  & 52.16 & 71.08  & 48.55   &    67.69   \\
DGMP, $\lambda=10^3$ & \textbf{54.81} & \textbf{73.94}  & \textbf{53.54} & \textbf{72.56}  & \textbf{49.37}   &  \textbf{68.83}      \\
DGMP, $\lambda=10^5$ & 52.41 & 71.22  & 48.98 & 68.00  & 45.24   &    64.47    \\
\bottomrule
\end{tabular}
\label{tab:writer-gmp-margin}
\end{table}

Global average pooling
and global max pooling serve as baselines for our experiments. Not only are 
these used in practice, these pooling operators also represent special cases of
the other generalized and learned pooling functions. We evaluate different margin
parameters for the triplet loss $m \in \{ 0.1, 0.2, 0.5\}$ and different initializations for the DGMP pooling parameter 
$\lambda \in \{1, 10^3, 10^5 \}$, afterwards $\lambda$ is optimized during training.
The results are summarized in \cref{tab:writer-gmp-margin}. 
Generalized max pooling performs almost always
better than average pooling and max pooling for a given margin. 
For DGMP, the initialization with $\lambda = 10^3$ achieves the overall best results.

The choice of the margin parameter influences all models. All our models show 
the best performance at $m = 0.1$, regardless of the pooling function. 
However, some pooling layers seem more sensitive to the change in margin than others.
While average pooling is less susceptible to a change in the triplet loss margin,
max pooling suffers the most, dropping from \SI{50.32}{\percent} \map to \SI{40.04}{\percent} \map. 
As a result, DGMP outperforms both average pooling and max pooling considerably for a margin of $m = 0.1$, for $m = 0.5$, however, the performance of DGMP relative to the average pooling baseline depends on the proper choice of~$\lambda$. %

\begin{table}[t]
\caption{Comparison with different global pooling methods 
(global average pooling (Avg), global max pooling (Max), mixed pooling (Mixed), log-sum-exp pooling (LSE), deep generalized max pooling (DGMP))
and state of the art using the \icdars dataset dataset. All methods marked by (*) are taken from~\cite{Fiel17ICDAR}.}
\centering
\begin{tabular}{lcc}
\toprule
            &  \map   & rank-1 \\ 
\midrule 
Avg         & 50.4 & 69.2  \\       
Max         & 50.3 & 69.6  \\              
Mixed       & 50.2 & 69.6  \\
LSE         & 51.4 & 70.0  \\
DGMP (ours) & \textbf{54.8} & \textbf{73.9}  \\  
\midrule
Triplet Network (Fribourg)$^*$    &   30.7      & 47.8        \\
CoHinge features (Groningen)$^*$ &   54.2    & 76.1        \\
oBIF (Tébessa)$^*$           &   55.6      & 76.4        \\
Unsupervised~\cite{Christlein17ICDAR} & \textbf{76.2} & \textbf{88.9}   \\
\bottomrule
\end{tabular}
\label{tab:writer-all-pooling}  
\end{table}
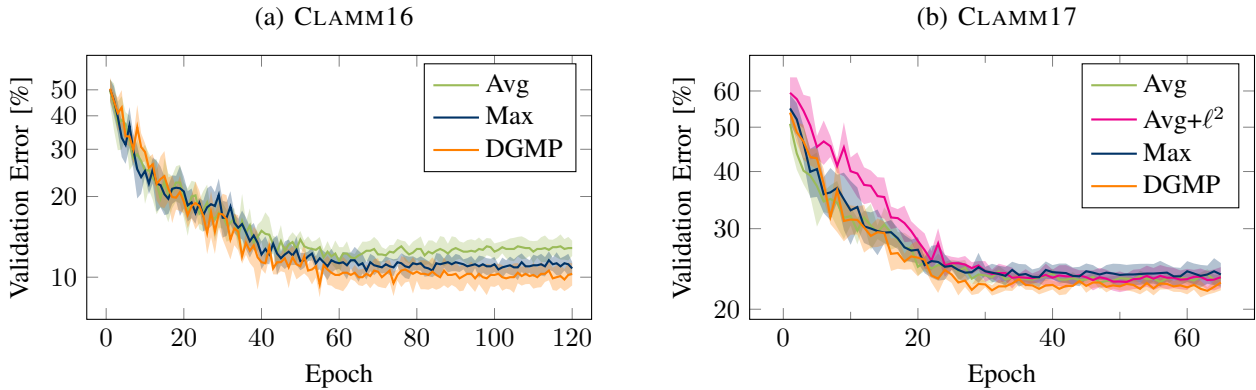
\begin{figure*}[t]
	\centering
	\begin{tikzpicture}
		\begin{axis}[scale only axis, 
				height=3.5cm, width=0.75\columnwidth,
				title={(a) \clammsix\phantomsubcaption\label{fig:clamm16_val}},
				xlabel={Epoch},
				ylabel={Validation Error [\%]},
				no markers,
				ymode=log,
				ytick={10,20,30,40,50},
				log ticks with fixed point,
				legend cell align={left},
				legend pos = north east,
				legend entries={Avg,Max,DGMP},
				xmax=125, xmin=-5,
				restrict x to domain={0:120},
			]
			\addlegendimage{no markers, ggreen, thick}
			\addlegendimage{no markers, faublue, thick}
			\addlegendimage{no markers, orange, thick}

			\myerrorplot{eval2/clamm16_avg_adam4e-4_pretrained_lcs1_nol2_mean_std.csv}{ggreen}
			\myerrorplot{eval2/clamm16_max_adam4e-4_pretrained_lcs1_nol2_mean_std.csv}{faublue}
			\myerrorplot{eval2/clamm16_gmp_adam4e-4_pretrained_lambda5k_lcs1_new_mean_std.csv}{orange}

			\addplot+[thick, ggreen, solid] table
				[x expr=(\coordindex+1), y expr={100-\thisrowno{0}*100}, col sep=comma]
			{%
				eval2/clamm16_avg_adam4e-4_pretrained_lcs1_nol2_mean_std.csv%
			};
			\addplot+[thick, faublue, solid] table
				[x expr=(\coordindex+1), y expr={100-\thisrowno{0}*100}, col sep=comma]
			{%
				eval2/clamm16_max_adam4e-4_pretrained_lcs1_nol2_mean_std.csv%
			};
			\addplot+[thick, orange, solid] table
				[x expr=(\coordindex+1), y expr={100-\thisrowno{0}*100}, col sep=comma]
			{%
				eval2/clamm16_gmp_adam4e-4_pretrained_lambda5k_lcs1_new_mean_std.csv%
			};
		\end{axis}
	\end{tikzpicture}
\qquad
	\begin{tikzpicture}
		\begin{axis}[scale only axis, 
				height=3.5cm, width=0.75\columnwidth,
				title={(b) \clammseven\phantomsubcaption\label{fig:clamm17_val}},
				xlabel={Epoch},
				ylabel={Validation Error [\%]},
				no markers,
				ymode=log,
				ytick={20,30,40,50,60},
				log ticks with fixed point,
				minor x tick num=1,
				legend cell align={left},
				legend pos = north east,
				legend entries={Avg,Avg+$\ell^2$,Max,DGMP},
				xmax=70, xmin=-5,
				restrict x to domain={0:65},
			]
			\addlegendimage{no markers, ggreen, thick}
			\addlegendimage{no markers, magenta, thick}
			\addlegendimage{no markers, faublue, thick}
			\addlegendimage{no markers, orange, thick}

			\myerrorplot{eval2/clamm17_avg_adam4e-4_pretrained_lcs1_final_mean_std.csv}{ggreen}
			\myerrorplot{eval2/clamm17_avg_adam4e-4_pretrained_lcs1_newnew_l2_mean_std.csv}{magenta}
			\myerrorplot{eval2/clamm17_max_adam4e-4_pretrained_lcs1_newnew_mean_std.csv}{faublue}
			\myerrorplot{eval2/clamm17_gmp_adam4e-4_pretrained_lambda5k_lcs1_ilr_mean_std.csv}{orange}

			\addplot+[thick, ggreen, solid] table
				[x expr=(\coordindex+1), y expr={100-\thisrowno{0}*100}, col sep=comma]
			{%
				eval2/clamm17_avg_adam4e-4_pretrained_lcs1_final_mean_std.csv%
			};
			\addplot+[thick, magenta, solid] table
				[x expr=(\coordindex+1), y expr={100-\thisrowno{0}*100}, col sep=comma]
			{%
				eval2/clamm17_avg_adam4e-4_pretrained_lcs1_newnew_l2_mean_std.csv%
			};

			\addplot+[thick, faublue, solid] table
				[x expr=(\coordindex+1), y expr={100-\thisrowno{0}*100}, col sep=comma]
			{%
				eval2/clamm17_max_adam4e-4_pretrained_lcs1_newnew_mean_std.csv%
			};
			\addplot+[thick, orange, solid] table
				[x expr=(\coordindex+1), y expr={100-\thisrowno{0}*100}, col sep=comma]
			{%
				eval2/clamm17_gmp_adam4e-4_pretrained_lambda5k_lcs1_ilr_mean_std.csv%
			};

		\end{axis}
	\end{tikzpicture}
		\caption{Mean validation error per epoch of the different pooling schemes computed of five runs with different initializations. The brighter area denotes the standard deviation. Please
		note the logarithmic y scale for improved clarity.}
	\label{fig:val_error}
\end{figure*}
Finally, we compare the proposed DGMP layer with other pooling methods, see \cref{tab:writer-all-pooling} (top) and the state of the art (bottom). 
Therefore, we also evaluated a weighted mixed pooling~\cite{Lee07} where a weighted sum of both pooling methods is computed and log-sum-exp pooling (LSE), \cf \cref{eq:mixed} and \cref{eq:lse}, respectively. 
The initial regularization parameters were set to $\alpha=0.5$ and $r=10$, which will get optimized during training.
All competing pooling techniques are clearly behind our proposed Deep GMP layer. When comparing with the state of the art, our method is inferior to the current state-of-the-art method of \etal{Christlein}~\cite{Christlein17ICDAR} where the features are learned in an unsupervised manner, which are explicitly aggregated. 
Yet, our method is superior to a competing deep learning method that uses the same triplet loss~\cite{Fiel17ICDAR}. 

\subsubsection{Script Type Classification}

Next, we want to see how our pooling layer performs for a classification task. 
Therefore, we classified images in twelve font classes. 
In contrast to writer identification, we evaluated them on the full images
at a resolution of $384\times384$. To compensate the lack of training data, we
made use of data augmentation. In addition to 
data standardization, we applied random resizing
(\SIrange{8}{100}{\percent}), random aspect ratios (0.75--1.33),   
random rotations of up to \SI{5}{\degree}, and random color
jitter of up to \SI{10}{\percent} brightness and contrast variations.

We modified the ResNet architecture slightly by reducing the stride of the last convolutional block to one (instead of two) to increase the spatial resolution before global pooling~\cite{Huang} resulting into activation maps of size $21\times21$. Thus, more activation vectors are pooled and the solution for the optimization problem becomes more reliable.
For optimization, we used the same setting as for writer identification, but 
we chose to use a higher learning rate ($\eta=4\cdot 10^{-4}$)
in the beginning. Though, the exponential decay was triggered earlier: starting
at epoch 30 for \clammsix and at epoch 16 for \clammseven, because both datasets are comparably small, and since we did not employ a patch-wise classification.
The regularization parameter ($\lambda$) was set to 5k a learning rate multiplier of 100 was used in the \clammseven experiment, which showed a slightly improved validation accuracy (no multiplier for the \clammsix experiment).

\Cref{fig:val_error} shows the mean and standard deviation (bright color) of the validation error against the epoch number for five runs.
In both cases, we see that DGMP achieves overall lower validation errors. Since
the validation set is quite small for the \clammsix dataset (204 samples), the
beginning of the validation curve of \cref{fig:clamm16_val} is quite wriggled.
Interestingly, max pooling has a lower validation error than average pooling.
For \cref{fig:clamm17_val}, the difference becomes clearer and we see that GMP
generalizes faster than max or average pooling. 

For the \clammseven dataset, we also experimented with another average pooling variant that enforces the pooled representation to have an $\ell^2$ norm of one. The experiments show that this needs a longer convergence time in comparison to normal global average pooling due to the bounded gradients, but results in about the same accuracy. 

\begin{table}[t]
	\caption{Classification results showing the average accuracy [\%] of five runs (Avg) and an ensemble (Ens) computed by all five models. Note that the state of the art approaches (bottom) use multiple crops while we only use a single view.}
	\label{tab:classification}
	\centering
\subcaptionbox{\clammsix\label{tab:clamm16}}{
	\begin{tabular}{l*{2}c}
	\toprule
	Pooling	& Avg & Ens\\
	\midrule
  Avg & 85.9 & 87.9\\
  Max &  85.5 & 86.8\\
  DGMP (ours) & \textbf{87.1} & \textbf{88.6}\\
\midrule	
	FAU~\cite{Kestemont17} 				& 83.9 & --\\
	ResNet50~\cite{Christlein18PHD} 	& 86.3 & -- \\
	ResNet50~\cite{Tensmeyer17} 		& --   & 86.6\\
	
\bottomrule
	\end{tabular}
}
\smallskip
\subcaptionbox{\clammseven\label{tab:clamm17}}{
	\begin{tabular}{l*{2}c}
	\toprule
	Pooling	& Avg & Ens\\
	\midrule
  Avg &  81.3 & 82.6\\
  Avg + $\ell^2$ & 81.2 & 82.2\\
  Max &  81.3 & 83.2 \\
  DGMP (ours) & \textbf{83.3} & 85.1\\
	\midrule
	CK2~\cite{Cloppet17} & 78.9 & -- \\
	T-DeepCNN~\cite{Cloppet17} & -- & \textbf{85.2}\\
	\bottomrule
	\end{tabular}
}
\end{table}
The average results of the five best models (obtained by taking for each run the model with lowest validation error) can be seen in \cref{tab:classification}. 
The standard deviations are: (a) \clammsix: \SI{1.4}{\percent},
\SI{0.5}{\percent}, and \SI{0.6}{\percent}; (b) \clammseven: 
\SI{0.4}{\percent}, \SI{0.8}{\percent}, and \SI{1}{\percent} 
for average pooling, max pooling, and DGMP, respectively.

Additionally, we provide the results when computing an ensemble of the five
models. In particular, we compute the predictions by averaging all activations of the
last linear layers. In advance, we make the outcomes more comparable by applying
the softmax function. Furthermore, each individual output is weighted by
the validation accuracy of the respective model.

Similar to the writer identification results, our proposed DGMP layer outperforms the other pooling layers in both \clamm datasets. For the \clammsix dataset, we even achieve a new highscore with a single model. Using an ensemble, we are substantially better than the approach of \etal{Tensmeyer}~\cite{Tensmeyer17}, who use an ensemble of two different models. For the \clammseven dataset, our method is comparable to the T-DeepCNN submission~\cite{Cloppet17}, where an ensemble of five models trained with different image sizes is computed.

\section{Conclusion}
\label{sec:conclusion}
We presented Deep Generalized Max Pooling (DGMP), a new pooling layer that balances on a per-sample basis frequent and rare activations. While common average and max pooling operate individually on each activation map, DGMP considers all locations in the activation volume as input while adding only one additional parameter to the network. 
We show that DGMP is superior to the commonly used global average or max pooling. 
In  writer identification/retrieval and font classification, it achieves higher accuracies than those global pooling techniques.
While not the focus of this work, we also outperform another method on the \clammsix dataset while performing comparable on the \clammseven dataset. 
For future work, we would like to improve our work on writer identification, \eg by integrating this layer in the work of \etal{Christlein}~\cite{Christlein17ICDAR}. We would also like to 
expand our evaluation on other scenarios, such as word spotting where the commonly-used PHOCNet~\cite{Sudholt16} uses a global max pooling layer. 

\bibliographystyle{IEEEtran}
{\footnotesize
\bibliography{thesis}
}
\end{document}